\documentclass{article}

\usepackage{arxiv}
\usepackage{graphicx, amsmath, pgfplots, amsfonts, setspace, hyperref}
\usepackage{algorithm}
\usepackage{algpseudocode}

\pgfplotsset{compat=1.17}

\title{An Optimization-Based Supervised Learning Algorithm for PXRD Phase Fraction Estimation}

\author{ \href{https://orcid.org/0000-0003-1729-559X}{\includegraphics[scale=0.06]{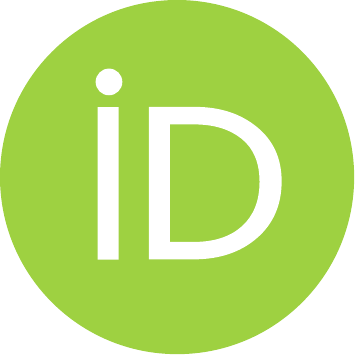}\hspace{1mm}Patrick Hosein}\\
	Department of Computer Science\\
	The University of the West Indies\\
	St. Augustine, Trinidad\\
	\texttt{patrick.hosein@sta.uwi.edu} \\
	\And
	\href{https://orcid.org/0000-0001-7821-6053}{\includegraphics[scale=0.06]{orcid.pdf}\hspace{1mm}Jaimie Greasley} \\
	Department of Physics\\
	The University of the West Indies\\
	St. Augustine, Trinidad\\
	\texttt{jaimie.greasley@gmail.com} \\
}

\renewcommand{\shorttitle}{Supervised Learning Algorithm for PXRD Phase Fraction Estimation}

\hypersetup{
pdftitle={An Optimization-Based Supervised Learning Algorithm for PXRD Phase Fraction Estimation},
pdfauthor={Patrick Hosein and Jamie Greasley},
colorlinks=true
}

\begin{document}

\maketitle

\begin{abstract}
In powder diffraction data analysis, phase identification is the process of determining the crystalline phases in a sample using its characteristic Bragg peaks. For multiphasic spectra, we must also determine the relative weight fraction of each phase in the sample. Machine Learning algorithms (e.g., Artificial Neural Networks) have been applied to perform such difficult tasks in powder diffraction analysis, but typically require a significant number of training samples for acceptable performance. We have developed an approach that performs well even with a small number of training samples. We apply a fixed-point iteration algorithm on the labelled training samples to estimate monophasic spectra. Then, given an unknown sample spectrum, we again use a fixed-point iteration algorithm to determine the weighted combination of monophase spectra that best approximates the unknown sample spectrum. These weights are the desired phase fractions for the sample. We compare our approach with several traditional Machine Learning algorithms. 

\end{abstract}

\section*{Main}
The assessment of powder X-ray diffraction (PXRD) spectra is central to many materials investigations. X-ray scattering data reveals important structural and micro-structural parameters for characterizing a material\cite{dinnebier2008powder}. The collection of scattered intensities attributed to labelled crystallographic planes, in fact serves as a fingerprint reference for the material structure\cite{pecharsky2009fundamentals}. 
Phase identification is performed by matching observed Bragg peaks to a powder pattern reference in a database, usually with the aid of a search-match program. Finding the corresponding reference may not be easy as different instrument settings or any slight deviation in structure cause variant diffraction profiles for any given phase.
For multi-phase analysis, phase fraction estimation is possible by presuming some relationship between the observed intensities for each phase in the spectrum. Several method varying in complexity, computational rigor and sample preparation requirements, are available \cite{zhou2018xrd}. These include the use of internal standard calibrations as in the Reference Intensity Ratio method \cite{hubbard1976reference}, or whole powder pattern fitting as with full-pattern summation \cite{smith1987quantitative, chipera2013fitting, butler2021powdr, m2021automated} and Rietveld refinement \cite{rietveld1967line}. 

Recently, Machine Learning has been used to characterize PXRD spectra \cite{agrawal2019deep, choudhary2022recent}. Some success has been reported for phase identification by both conventional Machine Learning models \cite{bunn2015generalized, bunn2016semi} as well as Deep Learning architectures \cite{lee2020deep, wang2020rapid, lee2021data, maffettone2021crystallography, szymanski2021probabilistic}. However, few studies have investigated phase quantification \cite{lee2020deep, lee2021data, park2022application}. 

Previously, we performed PXRD Rietveld characterization of mineral phases for a small batch of urinary stones \cite{greasley2022quantitative}. Rietveld refinement is a powerful, but time-consuming, pattern fitting procedure which employs least-squares minimization to obtain refined crystallographic parameters for a material, including phase fractions. While full crystallographic characterization may be useful for research in stone formation, it is not required for a clinical stone analysis program. Yet, estimated weight fractions serve to aid the analyst in differentiating between primary and secondary mineralization events. We therefore developed an approach, described in the next section, for estimating phase fractions given the spectrum of an unknown sample.

\subsection*{Proposed Algorithm}

We present the proposed model by first describing the training process and then showing how data derived from training is used for estimating phase-fractions for unknown samples. Note that this is a supervised Machine Learning approach, unlike full-pattern summation \cite{chipera2013fitting, butler2021powdr, m2021automated} which requires building a reference library prior to fitting. 

We used Leave-One-Out cross-validation by reserving one sample at a time for testing while all others are used for training.
Let $y_i(s)$ represent the intensity value for sample $s$ at angle $i$. Let $\alpha_j(s)$ represent the percentage of phase $j$ in sample $s$. Let $x_i(j)$ denote the reading that would be obtained at angle $i$ if the sample consisted solely of phase $j$. We denote the number of samples $N$, the number of phases $M$, and the number of angles $K$.

Consider the model whereby the intensity achieved for sample $s$ at angle $i$ consists of a weighted combination of the intensities obtained for the individual components $j$. As such the reading at angle $i$ for sample $s$ is approximated by
\begin{equation}
\hat{y}_i(s) = \sum_{j=1}^M \alpha_j(s) x_i(j)
\end{equation}
Note that in the training set, we know $\alpha_j(s)$ and $y_i(s)$ but we do not know $x_i(j)$ so we use the training set to obtain approximations to $x_i(j)$ that provides the lowest error. The sum squared error for the approximation for sample $s$ at angle $i$ is given by
\begin{equation}
E^2 = \sum_{i,s} (y_i(s) - \hat{y}_i(s))^2 = \sum_{i,s} \left( y_i(s) - \sum_{j=1}^M \alpha_j(s) x_i(j) \right)^2
\end{equation}
We then need to find values for $x_i(j)$ for all $i$ and $j$ that minimizes $E^2$. If we take the partial derivative of $E^2$ with respect to $x_i(j)$ we get
\begin{equation}
\frac{\partial E^2}{\partial x_i(j)} = \sum_{s} -2 \alpha_j(s) \left( y_i(s) - \sum_{j'=1}^M \alpha_{j'}(s) x_i(j') \right)
\end{equation}
and setting to zero we get
\begin{equation}
\sum_{s} \alpha_j(s) \left( y_i(s) - \sum_{j'=1}^M \alpha_{j'}(s) x_i(j') \right) = 0
\end{equation}

We use a coordinate gradient descent approach. For each variable, we take the derivative and update to the value that achieves a zero gradient. Note that this is a constrained optimization problem so if the zero gradient is achieved at a negative value then we instead set to zero.
\begin{equation}
x_i(j) \leftarrow \max \left\{ \frac{1}{\alpha_j(s)^2} \sum_{s} \alpha_j(s) \left( y_i(s) - \sum_{j'=1,j' \ne j}^M \alpha_{j'}(s) x_i(j') \right), 0 \right\}
\end{equation}

Next, we describe how testing was performed.
Once $x_i(j)$ is solved for all $i$ and $j$, these can be used to compute our estimate of $\alpha_j$ for a reserved test sample $s$. These estimated values are denoted as $\hat{\alpha}_j$. Knowing now $x_i(j)$ estimated from the training set, we have $y_i(s)$. We now need to find $\alpha_j(s)$. We let:
\[
\hat{y}_i(s) = \sum_{j=1}^M \hat{\alpha}_j(s) x_i(j)
\]
The sum squared error is again given by 
\[
E^2 = \sum_{i,s} \left( y_i(s) - \sum_{j=1}^M \alpha_j(s) x_i(j) \right)^2
\]
However, the partial derivative is now taken with respect to $\alpha_j(s)$ giving,
\[
\frac{\partial E^2}{\partial \alpha_j(s)} = \sum_{i} -2 x_i(j) \left( y_i(s) - \sum_{j'=1}^M \alpha_{j'}(s) x_i(j') \right)
\]
and by setting to zero:
\[
 \sum_{i} x_i(j) \left( y_i(s) - \sum_{j'=1}^M \alpha_{j'}(s) x_i(j') \right) = 0
\]
We use again a coordinate gradient descent approach. For each variable, we take the derivative and update to the value that achieves a zero gradient. If the zero gradient is achieved at a negative value, then that is instead set to zero.
\begin{equation}
\alpha_j(s) \leftarrow \max \left\{ \frac{1}{x_i(j)^2} \sum_i x_i(j) \left( y_i(s) - \sum_{j'=1,j' \ne j}^M \alpha_{j'}(s) x_i(j') \right), 0 \right\}
\end{equation}
The components of $\alpha_j(s)$ must sum to one, hence the following scaling is performed after completing each iteration. 
\begin{equation}
\alpha_j(s) \leftarrow \frac{\alpha_j(s)}{\sum_j \alpha_j(s)}
\end{equation}
These steps are repeated until convergence is achieved, typically after 15 iterations.

Let us now introduce a suitable performance metric.
Consider a test sample, the vector of true phase fractions is denoted by $\alpha$ and the vector of predicted fractions by $\hat{\alpha}$. We write the $||\alpha - \hat{\alpha}||_2$ to represent the Euclidean distance between the vectors. The sum of the elements in each of the vectors $\alpha$ and $\hat{\alpha}$ is 1, hence one can show that the maximum value of this distance is $\sqrt{2}$. We define our performance metric $\rho$ as
\begin{equation}
\rho \equiv 1 - \frac{||\alpha - \hat{\alpha}||_2}{\sqrt{2}}
\end{equation}

If therefore $\alpha = \hat{\alpha}$, then $\rho=1$ implying a perfect prediction. Supposing there is actually only one phase, and the model predicts one phase as well, but that the predicted is different to the actual, then $\rho=0$ and the prediction is completely wrong. Consider a bi-phasic sample where the two phases are correctly predicted. However, if the true fractions are 0.5 and 0.5 and the predicted fractions are 0.6 and 0.4, then $\rho=0.9$ indicating a close but not perfect prediction.

First, all samples were used for training and the predictions computed for each with the above algorithm. This step is equivalent to evaluating the performance of the training set, for which we obtained $\rho = 0.89$. Next, Leave-Out-One cross-validation was carried out such that one sample at a time was reserved for testing while all others were used in training. Here, we obtained $\rho = 0.86$, finding that the model did not over-fit as the testing and training performances are similar. What more is important is the closeness in match of the prediction to the actual for the individual phases. Figure \ref{bar} depicts bar charts indicating phase fractions for the 7 possible phases for the 46 samples with LOOCV. Note that the model always correctly predicted the dominant phase even if the actual fractional value did not match perfectly.

\pgfplotsset{ybar, height=2.5cm, width=0.55\textwidth, yticklabels=\empty, xticklabels=\empty, ytick=\empty, xtick=\empty, ymin=0, ymax=1}
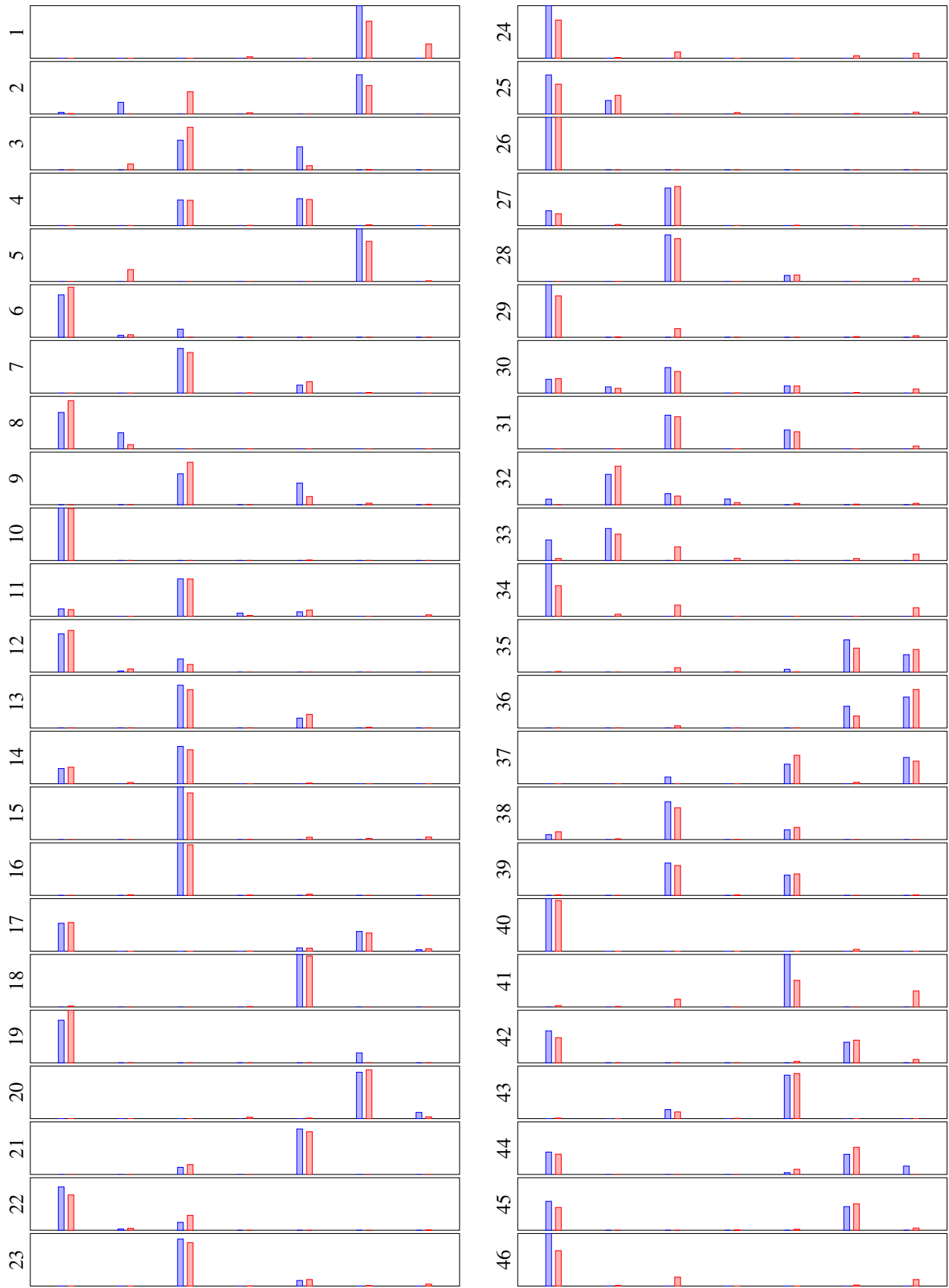
\begin{figure}[ht]
\begin{tikzpicture}
\begin{axis}[name=1, bar width=0.1cm, ylabel=1]
\addplot coordinates { (1, 0.0 ) (2, 0.0 ) (3, 0.0 ) (4, 0.0 ) (5, 0.0 ) (6, 1.0 ) (7, 0.0 )};
\addplot coordinates { (1, 0.0 ) (2, 0.0 ) (3, 0.0 ) (4, 0.026 ) (5, 0.0 ) (6, 0.704 ) (7, 0.27 )};
\end{axis}
\begin{axis}[name=2, at={(1.below south west)}, yshift=-0.05cm, anchor=north west, bar width=0.1cm, ylabel=2]
\addplot coordinates { (1, 0.031 ) (2, 0.222 ) (3, 0.0 ) (4, 0.0 ) (5, 0.0 ) (6, 0.747 ) (7, 0.0 )};
\addplot coordinates { (1, 0.01 ) (2, 0.0 ) (3, 0.423 ) (4, 0.025 ) (5, 0.0 ) (6, 0.542 ) (7, 0.0 )};
\end{axis}
\begin{axis}[name=3, at={(2.below south west)}, yshift=-0.05cm, anchor=north west, bar width=0.1cm, ylabel=3]
\addplot coordinates { (1, 0.0 ) (2, 0.0 ) (3, 0.563 ) (4, 0.0 ) (5, 0.437 ) (6, 0.0 ) (7, 0.0 )};
\addplot coordinates { (1, 0.0 ) (2, 0.111 ) (3, 0.81 ) (4, 0.0 ) (5, 0.078 ) (6, 0.001 ) (7, 0.0 )};
\end{axis}
\begin{axis}[name=4, at={(3.below south west)}, yshift=-0.05cm, anchor=north west, bar width=0.1cm, ylabel=4]
\addplot coordinates { (1, 0.0 ) (2, 0.0 ) (3, 0.487 ) (4, 0.0 ) (5, 0.513 ) (6, 0.0 ) (7, 0.0 )};
\addplot coordinates { (1, 0.0 ) (2, 0.0 ) (3, 0.483 ) (4, 0.002 ) (5, 0.497 ) (6, 0.018 ) (7, 0.0 )};
\end{axis}
\begin{axis}[name=5, at={(4.below south west)}, yshift=-0.05cm, anchor=north west, bar width=0.1cm, ylabel=5]
\addplot coordinates { (1, 0.0 ) (2, 0.0 ) (3, 0.0 ) (4, 0.0 ) (5, 0.0 ) (6, 1.0 ) (7, 0.0 )};
\addplot coordinates { (1, 0.0 ) (2, 0.226 ) (3, 0.0 ) (4, 0.0 ) (5, 0.0 ) (6, 0.762 ) (7, 0.012 )};
\end{axis}
\begin{axis}[name=6, at={(5.below south west)}, yshift=-0.05cm, anchor=north west, bar width=0.1cm, ylabel=6]
\addplot coordinates { (1, 0.808 ) (2, 0.038 ) (3, 0.154 ) (4, 0.0 ) (5, 0.0 ) (6, 0.0 ) (7, 0.0 )};
\addplot coordinates { (1, 0.953 ) (2, 0.047 ) (3, 0.0 ) (4, 0.0 ) (5, 0.0 ) (6, 0.0 ) (7, 0.0 )};
\end{axis}
\begin{axis}[name=7, at={(6.below south west)}, yshift=-0.05cm, anchor=north west, bar width=0.1cm, ylabel=7]
\addplot coordinates { (1, 0.0 ) (2, 0.0 ) (3, 0.849 ) (4, 0.0 ) (5, 0.151 ) (6, 0.0 ) (7, 0.0 )};
\addplot coordinates { (1, 0.0 ) (2, 0.0 ) (3, 0.772 ) (4, 0.002 ) (5, 0.216 ) (6, 0.01 ) (7, 0.0 )};
\end{axis}
\begin{axis}[name=8, at={(7.below south west)}, yshift=-0.05cm, anchor=north west, bar width=0.1cm, ylabel=8]
\addplot coordinates { (1, 0.693 ) (2, 0.307 ) (3, 0.0 ) (4, 0.0 ) (5, 0.0 ) (6, 0.0 ) (7, 0.0 )};
\addplot coordinates { (1, 0.916 ) (2, 0.08 ) (3, 0.0 ) (4, 0.004 ) (5, 0.0 ) (6, 0.0 ) (7, 0.0 )};
\end{axis}
\begin{axis}[name=9, at={(8.below south west)}, yshift=-0.05cm, anchor=north west, bar width=0.1cm, ylabel=9]
\addplot coordinates { (1, 0.0 ) (2, 0.0 ) (3, 0.589 ) (4, 0.0 ) (5, 0.411 ) (6, 0.0 ) (7, 0.0 )};
\addplot coordinates { (1, 0.0 ) (2, 0.0 ) (3, 0.808 ) (4, 0.0 ) (5, 0.155 ) (6, 0.029 ) (7, 0.009 )};
\end{axis}
\begin{axis}[name=10, at={(9.below south west)}, yshift=-0.05cm, anchor=north west, bar width=0.1cm, ylabel=10]
\addplot coordinates { (1, 1.0 ) (2, 0.0 ) (3, 0.0 ) (4, 0.0 ) (5, 0.0 ) (6, 0.0 ) (7, 0.0 )};
\addplot coordinates { (1, 0.991 ) (2, 0.0 ) (3, 0.0 ) (4, 0.0 ) (5, 0.009 ) (6, 0.0 ) (7, 0.0 )};
\end{axis}
\begin{axis}[name=11, at={(10.below south west)}, yshift=-0.05cm, anchor=north west, bar width=0.1cm, ylabel=11]
\addplot coordinates { (1, 0.141 ) (2, 0.0 ) (3, 0.713 ) (4, 0.061 ) (5, 0.085 ) (6, 0.0 ) (7, 0.0 )};
\addplot coordinates { (1, 0.127 ) (2, 0.0 ) (3, 0.71 ) (4, 0.016 ) (5, 0.119 ) (6, 0.0 ) (7, 0.028 )};
\end{axis}
\begin{axis}[name=12, at={(11.below south west)}, yshift=-0.05cm, anchor=north west, bar width=0.1cm, ylabel=12]
\addplot coordinates { (1, 0.728 ) (2, 0.022 ) (3, 0.25 ) (4, 0.0 ) (5, 0.0 ) (6, 0.0 ) (7, 0.0 )};
\addplot coordinates { (1, 0.794 ) (2, 0.06 ) (3, 0.146 ) (4, 0.0 ) (5, 0.0 ) (6, 0.0 ) (7, 0.0 )};
\end{axis}
\begin{axis}[name=13, at={(12.below south west)}, yshift=-0.05cm, anchor=north west, bar width=0.1cm, ylabel=13]
\addplot coordinates { (1, 0.0 ) (2, 0.0 ) (3, 0.813 ) (4, 0.0 ) (5, 0.187 ) (6, 0.0 ) (7, 0.0 )};
\addplot coordinates { (1, 0.0 ) (2, 0.0 ) (3, 0.729 ) (4, 0.0 ) (5, 0.258 ) (6, 0.013 ) (7, 0.0 )};
\end{axis}
\begin{axis}[name=14, at={(13.below south west)}, yshift=-0.05cm, anchor=north west, bar width=0.1cm, ylabel=14]
\addplot coordinates { (1, 0.29 ) (2, 0.0 ) (3, 0.71 ) (4, 0.0 ) (5, 0.0 ) (6, 0.0 ) (7, 0.0 )};
\addplot coordinates { (1, 0.315 ) (2, 0.024 ) (3, 0.647 ) (4, 0.0 ) (5, 0.014 ) (6, 0.0 ) (7, 0.0 )};
\end{axis}
\begin{axis}[name=15, at={(14.below south west)}, yshift=-0.05cm, anchor=north west, bar width=0.1cm, ylabel=15]
\addplot coordinates { (1, 0.0 ) (2, 0.0 ) (3, 1.0 ) (4, 0.0 ) (5, 0.0 ) (6, 0.0 ) (7, 0.0 )};
\addplot coordinates { (1, 0.0 ) (2, 0.0 ) (3, 0.886 ) (4, 0.003 ) (5, 0.044 ) (6, 0.02 ) (7, 0.048 )};
\end{axis}
\begin{axis}[name=16, at={(15.below south west)}, yshift=-0.05cm, anchor=north west, bar width=0.1cm, ylabel=16]
\addplot coordinates { (1, 0.0 ) (2, 0.0 ) (3, 1.0 ) (4, 0.0 ) (5, 0.0 ) (6, 0.0 ) (7, 0.0 )};
\addplot coordinates { (1, 0.0 ) (2, 0.012 ) (3, 0.962 ) (4, 0.003 ) (5, 0.024 ) (6, 0.0 ) (7, 0.0 )};
\end{axis}
\begin{axis}[name=17, at={(16.below south west)}, yshift=-0.05cm, anchor=north west, bar width=0.1cm, ylabel=17]
\addplot coordinates { (1, 0.531 ) (2, 0.0 ) (3, 0.0 ) (4, 0.0 ) (5, 0.063 ) (6, 0.376 ) (7, 0.029 )};
\addplot coordinates { (1, 0.546 ) (2, 0.0 ) (3, 0.0 ) (4, 0.003 ) (5, 0.057 ) (6, 0.346 ) (7, 0.048 )};
\end{axis}
\begin{axis}[name=18, at={(17.below south west)}, yshift=-0.05cm, anchor=north west, bar width=0.1cm, ylabel=18]
\addplot coordinates { (1, 0.0 ) (2, 0.0 ) (3, 0.0 ) (4, 0.0 ) (5, 1.0 ) (6, 0.0 ) (7, 0.0 )};
\addplot coordinates { (1, 0.019 ) (2, 0.0 ) (3, 0.0 ) (4, 0.01 ) (5, 0.971 ) (6, 0.0 ) (7, 0.0 )};
\end{axis}
\begin{axis}[name=19, at={(18.below south west)}, yshift=-0.05cm, anchor=north west, bar width=0.1cm, ylabel=19]
\addplot coordinates { (1, 0.811 ) (2, 0.0 ) (3, 0.0 ) (4, 0.0 ) (5, 0.0 ) (6, 0.189 ) (7, 0.0 )};
\addplot coordinates { (1, 1.0 ) (2, 0.0 ) (3, 0.0 ) (4, 0.0 ) (5, 0.0 ) (6, 0.0 ) (7, 0.0 )};
\end{axis}
\begin{axis}[name=20, at={(19.below south west)}, yshift=-0.05cm, anchor=north west, bar width=0.1cm, ylabel=20]
\addplot coordinates { (1, 0.0 ) (2, 0.0 ) (3, 0.0 ) (4, 0.0 ) (5, 0.0 ) (6, 0.882 ) (7, 0.118 )};
\addplot coordinates { (1, 0.0 ) (2, 0.0 ) (3, 0.0 ) (4, 0.027 ) (5, 0.01 ) (6, 0.928 ) (7, 0.035 )};
\end{axis}
\begin{axis}[name=21, at={(20.below south west)}, yshift=-0.05cm, anchor=north west, bar width=0.1cm, ylabel=21]
\addplot coordinates { (1, 0.0 ) (2, 0.0 ) (3, 0.135 ) (4, 0.0 ) (5, 0.865 ) (6, 0.0 ) (7, 0.0 )};
\addplot coordinates { (1, 0.0 ) (2, 0.0 ) (3, 0.188 ) (4, 0.0 ) (5, 0.812 ) (6, 0.0 ) (7, 0.0 )};
\end{axis}
\begin{axis}[name=22, at={(21.below south west)}, yshift=-0.05cm, anchor=north west, bar width=0.1cm, ylabel=22]
\addplot coordinates { (1, 0.826 ) (2, 0.025 ) (3, 0.15 ) (4, 0.0 ) (5, 0.0 ) (6, 0.0 ) (7, 0.0 )};
\addplot coordinates { (1, 0.672 ) (2, 0.037 ) (3, 0.283 ) (4, 0.0 ) (5, 0.0 ) (6, 0.0 ) (7, 0.007 )};
\end{axis}
\begin{axis}[name=23, at={(22.below south west)}, yshift=-0.05cm, anchor=north west, bar width=0.1cm, ylabel=23]
\addplot coordinates { (1, 0.0 ) (2, 0.0 ) (3, 0.894 ) (4, 0.0 ) (5, 0.106 ) (6, 0.0 ) (7, 0.0 )};
\addplot coordinates { (1, 0.0 ) (2, 0.0 ) (3, 0.826 ) (4, 0.003 ) (5, 0.124 ) (6, 0.009 ) (7, 0.038 )};
\end{axis}
\end{tikzpicture}
\hspace{10pt}
\begin{tikzpicture}
\begin{axis}[name=24, bar width=0.1cm, ylabel=24]
\addplot coordinates { (1, 1.0 ) (2, 0.0 ) (3, 0.0 ) (4, 0.0 ) (5, 0.0 ) (6, 0.0 ) (7, 0.0 )};
\addplot coordinates { (1, 0.726 ) (2, 0.013 ) (3, 0.12 ) (4, 0.0 ) (5, 0.0 ) (6, 0.048 ) (7, 0.093 )};
\end{axis}
\begin{axis}[name=25, at={(24.below south west)}, yshift=-0.05cm, anchor=north west, bar width=0.1cm, ylabel=25]
\addplot coordinates { (1, 0.743 ) (2, 0.257 ) (3, 0.0 ) (4, 0.0 ) (5, 0.0 ) (6, 0.0 ) (7, 0.0 )};
\addplot coordinates { (1, 0.569 ) (2, 0.355 ) (3, 0.0 ) (4, 0.028 ) (5, 0.0 ) (6, 0.012 ) (7, 0.036 )};
\end{axis}
\begin{axis}[name=26, at={(25.below south west)}, yshift=-0.05cm, anchor=north west, bar width=0.1cm, ylabel=26]
\addplot coordinates { (1, 1.0 ) (2, 0.0 ) (3, 0.0 ) (4, 0.0 ) (5, 0.0 ) (6, 0.0 ) (7, 0.0 )};
\addplot coordinates { (1, 1.0 ) (2, 0.0 ) (3, 0.0 ) (4, 0.0 ) (5, 0.0 ) (6, 0.0 ) (7, 0.0 )};
\end{axis}
\begin{axis}[name=27, at={(26.below south west)}, yshift=-0.05cm, anchor=north west, bar width=0.1cm, ylabel=27]
\addplot coordinates { (1, 0.284 ) (2, 0.0 ) (3, 0.716 ) (4, 0.0 ) (5, 0.0 ) (6, 0.0 ) (7, 0.0 )};
\addplot coordinates { (1, 0.225 ) (2, 0.023 ) (3, 0.745 ) (4, 0.0 ) (5, 0.006 ) (6, 0.0 ) (7, 0.0 )};
\end{axis}
\begin{axis}[name=28, at={(27.below south west)}, yshift=-0.05cm, anchor=north west, bar width=0.1cm, ylabel=28]
\addplot coordinates { (1, 0.0 ) (2, 0.0 ) (3, 0.885 ) (4, 0.0 ) (5, 0.115 ) (6, 0.0 ) (7, 0.0 )};
\addplot coordinates { (1, 0.0 ) (2, 0.003 ) (3, 0.815 ) (4, 0.003 ) (5, 0.123 ) (6, 0.0 ) (7, 0.056 )};
\end{axis}
\begin{axis}[name=29, at={(28.below south west)}, yshift=-0.05cm, anchor=north west, bar width=0.1cm, ylabel=29]
\addplot coordinates { (1, 1.0 ) (2, 0.0 ) (3, 0.0 ) (4, 0.0 ) (5, 0.0 ) (6, 0.0 ) (7, 0.0 )};
\addplot coordinates { (1, 0.787 ) (2, 0.005 ) (3, 0.164 ) (4, 0.0 ) (5, 0.0 ) (6, 0.013 ) (7, 0.03 )};
\end{axis}
\begin{axis}[name=30, at={(29.below south west)}, yshift=-0.05cm, anchor=north west, bar width=0.1cm, ylabel=30]
\addplot coordinates { (1, 0.262 ) (2, 0.118 ) (3, 0.485 ) (4, 0.0 ) (5, 0.136 ) (6, 0.0 ) (7, 0.0 )};
\addplot coordinates { (1, 0.274 ) (2, 0.09 ) (3, 0.41 ) (4, 0.001 ) (5, 0.135 ) (6, 0.012 ) (7, 0.077 )};
\end{axis}
\begin{axis}[name=31, at={(30.below south west)}, yshift=-0.05cm, anchor=north west, bar width=0.1cm, ylabel=31]
\addplot coordinates { (1, 0.0 ) (2, 0.0 ) (3, 0.641 ) (4, 0.0 ) (5, 0.359 ) (6, 0.0 ) (7, 0.0 )};
\addplot coordinates { (1, 0.0 ) (2, 0.005 ) (3, 0.611 ) (4, 0.003 ) (5, 0.327 ) (6, 0.0 ) (7, 0.054 )};
\end{axis}
\begin{axis}[name=32, at={(31.below south west)}, yshift=-0.05cm, anchor=north west, bar width=0.1cm, ylabel=32]
\addplot coordinates { (1, 0.105 ) (2, 0.577 ) (3, 0.21 ) (4, 0.108 ) (5, 0.0 ) (6, 0.0 ) (7, 0.0 )};
\addplot coordinates { (1, 0.0 ) (2, 0.734 ) (3, 0.164 ) (4, 0.039 ) (5, 0.025 ) (6, 0.011 ) (7, 0.026 )};
\end{axis}
\begin{axis}[name=33, at={(32.below south west)}, yshift=-0.05cm, anchor=north west, bar width=0.1cm, ylabel=33]
\addplot coordinates { (1, 0.391 ) (2, 0.609 ) (3, 0.0 ) (4, 0.0 ) (5, 0.0 ) (6, 0.0 ) (7, 0.0 )};
\addplot coordinates { (1, 0.038 ) (2, 0.502 ) (3, 0.26 ) (4, 0.042 ) (5, 0.0 ) (6, 0.038 ) (7, 0.12 )};
\end{axis}
\begin{axis}[name=34, at={(33.below south west)}, yshift=-0.05cm, anchor=north west, bar width=0.1cm, ylabel=34]
\addplot coordinates { (1, 1.0 ) (2, 0.0 ) (3, 0.0 ) (4, 0.0 ) (5, 0.0 ) (6, 0.0 ) (7, 0.0 )};
\addplot coordinates { (1, 0.584 ) (2, 0.04 ) (3, 0.213 ) (4, 0.0 ) (5, 0.0 ) (6, 0.0 ) (7, 0.163 )};
\end{axis}
\begin{axis}[name=35, at={(34.below south west)}, yshift=-0.05cm, anchor=north west, bar width=0.1cm, ylabel=35]
\addplot coordinates { (1, 0.0 ) (2, 0.0 ) (3, 0.0 ) (4, 0.0 ) (5, 0.054 ) (6, 0.614 ) (7, 0.332 )};
\addplot coordinates { (1, 0.012 ) (2, 0.0 ) (3, 0.085 ) (4, 0.009 ) (5, 0.006 ) (6, 0.457 ) (7, 0.431 )};
\end{axis}
\begin{axis}[name=36, at={(35.below south west)}, yshift=-0.05cm, anchor=north west, bar width=0.1cm, ylabel=36]
\addplot coordinates { (1, 0.0 ) (2, 0.0 ) (3, 0.0 ) (4, 0.0 ) (5, 0.0 ) (6, 0.413 ) (7, 0.587 )};
\addplot coordinates { (1, 0.0 ) (2, 0.0 ) (3, 0.041 ) (4, 0.0 ) (5, 0.0 ) (6, 0.227 ) (7, 0.732 )};
\end{axis}
\begin{axis}[name=37, at={(36.below south west)}, yshift=-0.05cm, anchor=north west, bar width=0.1cm, ylabel=37]
\addplot coordinates { (1, 0.0 ) (2, 0.0 ) (3, 0.127 ) (4, 0.0 ) (5, 0.373 ) (6, 0.0 ) (7, 0.5 )};
\addplot coordinates { (1, 0.0 ) (2, 0.0 ) (3, 0.0 ) (4, 0.0 ) (5, 0.542 ) (6, 0.028 ) (7, 0.43 )};
\end{axis}
\begin{axis}[name=38, at={(37.below south west)}, yshift=-0.05cm, anchor=north west, bar width=0.1cm, ylabel=38]
\addplot coordinates { (1, 0.093 ) (2, 0.0 ) (3, 0.72 ) (4, 0.0 ) (5, 0.187 ) (6, 0.0 ) (7, 0.0 )};
\addplot coordinates { (1, 0.147 ) (2, 0.014 ) (3, 0.606 ) (4, 0.001 ) (5, 0.231 ) (6, 0.0 ) (7, 0.0 )};
\end{axis}
\begin{axis}[name=39, at={(38.below south west)}, yshift=-0.05cm, anchor=north west, bar width=0.1cm, ylabel=39]
\addplot coordinates { (1, 0.0 ) (2, 0.0 ) (3, 0.615 ) (4, 0.0 ) (5, 0.385 ) (6, 0.0 ) (7, 0.0 )};
\addplot coordinates { (1, 0.01 ) (2, 0.0 ) (3, 0.568 ) (4, 0.006 ) (5, 0.408 ) (6, 0.0 ) (7, 0.009 )};
\end{axis}
\begin{axis}[name=40, at={(39.below south west)}, yshift=-0.05cm, anchor=north west, bar width=0.1cm, ylabel=40]
\addplot coordinates { (1, 1.0 ) (2, 0.0 ) (3, 0.0 ) (4, 0.0 ) (5, 0.0 ) (6, 0.0 ) (7, 0.0 )};
\addplot coordinates { (1, 0.965 ) (2, 0.0 ) (3, 0.0 ) (4, 0.0 ) (5, 0.0 ) (6, 0.035 ) (7, 0.0 )};
\end{axis}
\begin{axis}[name=41, at={(40.below south west)}, yshift=-0.05cm, anchor=north west, bar width=0.1cm, ylabel=41]
\addplot coordinates { (1, 0.0 ) (2, 0.0 ) (3, 0.0 ) (4, 0.0 ) (5, 1.0 ) (6, 0.0 ) (7, 0.0 )};
\addplot coordinates { (1, 0.027 ) (2, 0.011 ) (3, 0.147 ) (4, 0.003 ) (5, 0.507 ) (6, 0.0 ) (7, 0.305 )};
\end{axis}
\begin{axis}[name=42, at={(41.below south west)}, yshift=-0.05cm, anchor=north west, bar width=0.1cm, ylabel=42]
\addplot coordinates { (1, 0.608 ) (2, 0.0 ) (3, 0.0 ) (4, 0.0 ) (5, 0.0 ) (6, 0.392 ) (7, 0.0 )};
\addplot coordinates { (1, 0.478 ) (2, 0.0 ) (3, 0.0 ) (4, 0.0 ) (5, 0.028 ) (6, 0.431 ) (7, 0.063 )};
\end{axis}
\begin{axis}[name=43, at={(42.below south west)}, yshift=-0.05cm, anchor=north west, bar width=0.1cm, ylabel=43]
\addplot coordinates { (1, 0.0 ) (2, 0.0 ) (3, 0.173 ) (4, 0.0 ) (5, 0.827 ) (6, 0.0 ) (7, 0.0 )};
\addplot coordinates { (1, 0.01 ) (2, 0.0 ) (3, 0.127 ) (4, 0.005 ) (5, 0.858 ) (6, 0.0 ) (7, 0.0 )};
\end{axis}
\begin{axis}[name=44, at={(43.below south west)}, yshift=-0.05cm, anchor=north west, bar width=0.1cm, ylabel=44]
\addplot coordinates { (1, 0.425 ) (2, 0.0 ) (3, 0.0 ) (4, 0.0 ) (5, 0.033 ) (6, 0.383 ) (7, 0.159 )};
\addplot coordinates { (1, 0.386 ) (2, 0.0 ) (3, 0.0 ) (4, 0.0 ) (5, 0.098 ) (6, 0.516 ) (7, 0.0 )};
\end{axis}
\begin{axis}[name=45, at={(44.below south west)}, yshift=-0.05cm, anchor=north west, bar width=0.1cm, ylabel=45]
\addplot coordinates { (1, 0.55 ) (2, 0.0 ) (3, 0.0 ) (4, 0.0 ) (5, 0.0 ) (6, 0.45 ) (7, 0.0 )};
\addplot coordinates { (1, 0.434 ) (2, 0.0 ) (3, 0.0 ) (4, 0.003 ) (5, 0.018 ) (6, 0.504 ) (7, 0.042 )};
\end{axis}
\begin{axis}[name=46, at={(45.below south west)}, yshift=-0.05cm, anchor=north west, bar width=0.1cm, ylabel=46]
\addplot coordinates { (1, 1.0 ) (2, 0.0 ) (3, 0.0 ) (4, 0.0 ) (5, 0.0 ) (6, 0.0 ) (7, 0.0 )};
\addplot coordinates { (1, 0.673 ) (2, 0.012 ) (3, 0.17 ) (4, 0.0 ) (5, 0.0 ) (6, 0.02 ) (7, 0.124 )};
\end{axis}
\end{tikzpicture}
\caption{Phase Fraction Predictions of Proposed Model with LOOCV. Predicted weight fractions are red bars, where the Rietveld-refined values are the blue bars.}
\label{bar}
\end{figure}

\begin{figure}[ht]
\centering
\includegraphics[width=\textwidth]{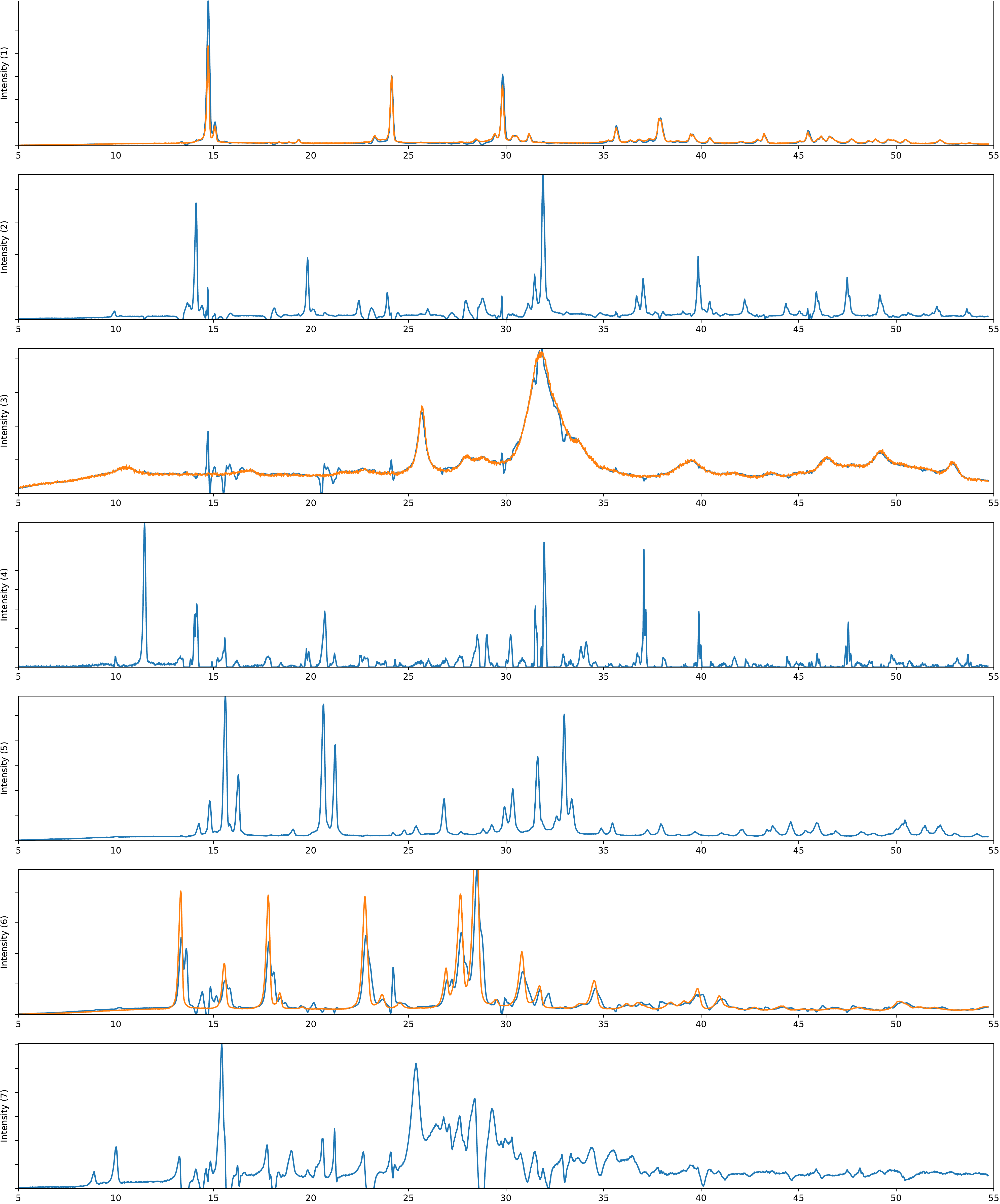}
\caption{Model-Estimated Intensity Spectra for Phase 1-7 (blue) with Overlay of Actual Mono-phase Experimental Examples (orange).}
\label{single}
\end{figure}

Figure \ref{single} visualizes the estimated $x(j)$ intensities versus the angle $i$, illustrating how closely this approach is able to estimate single-phase spectra. We provide an overlay of the spectra of actual monophase samples in orange.

\subsection*{Performance Comparison}

We have compared the proposed algorithm with other traditional Machine Learning models. In addition to the $\rho$ metric described in the previous section, we include two additional metrics, Mean Absolute Error (MAE) and Cosine Similarity (CS) defined as:
\begin{equation}
    {\sf MAE}=\frac{1}{MK} \sum_{s,j} \left|\alpha_{s,j}-\hat{\alpha}_{s,j}\right| \quad\quad
     {\sf CS} \equiv \frac{\alpha\cdot\hat{\alpha}}{||\alpha||\times||\hat{\alpha}||}
    \label{eq:mae}
\end{equation}

 We investigated several regression algorithms including K-Nearest Neighbours (KNN),  Support Vector Machine (SVM), Decision Tree (DT), Random Forest (RF), Extremely Randomized Trees (ERT) and Artificial Neural Network (ANN). Note that in order to get the best results for each model, significant additional work was necessary, e.g.  hyper-parameter tuning. This is not required for our proposed algorithm. The averaged performance results are summarized in Table \ref{tab:regress}. 

\begin{table}[ht]
\centering
\setlength{\tabcolsep}{15pt}
\renewcommand{\arraystretch}{1.5}
\caption{Leave-One-Out Cross-Validation Performance Results for Phase Fraction Estimation.}
\label{tab:regress}
\begin{tabular}{|c|r|r|r|r|r|r|r|} \hline
 & KNN & SVM & DT & RF & ERT & ANN & Proposed \\ \hline
\textbf{$\rho$} & 0.8369 & 0.7971 & 0.7438 & 0.7908 & 0.8217 & 0.8582 &{\bf 0.8641} \\ \hline
\textbf{MAE} & 0.0520 & 0.0773 & 0.0832 & 0.0778 & 0.0635 & 0.0491 & {\bf 0.0454} \\ \hline
\textbf{CS} & 0.9470 & 0.9364 & 0.8211 & 0.9029 & 0.9232 & {\bf 0.9662} & 0.9564  \\ \hline
\end{tabular}
\end{table}

The proposed model performed best in $\rho$ and MAE metrics but not for CS, where the ANN obtained a higher score. However, we noticed the ANN and SVM regressors did not strictly output positive, normalized weight fractions as the KNN and tree-based regressors did. The tabled results follow the performance after we applied these conditions to their outputs. Prior to this, the SVM gave $\text{MAE}=0.0913$ and $\text{CS}=0.9268$, and the ANN gave $\text{MAE}=0.0594$ and $\text{CS}=0.9627$.

Regarding phase fraction prediction, Lee et al.\cite{lee2021data} trained deep convolutional neural networks with millions of synthetic data relating to solid-state electrolytes, but achieved the best results with the SVM followed by RF, KNN and ANN models. For their application in gas-hydrate bearing sediments, Park et al.\cite{park2022application} obtained the best performance with RF amidst testing other neural networks architectures.

In this work, we have presented a new supervised learning approach for estimating phase fractions and illustrated its superiority by comparing with other ML models on our dataset. The assessment of kidney stones is however only one of very many possible contexts to which the approach may be applied. The model computes non-negative fractions, does not require parameter tuning, does not suffer from scaling issues as with other ML algorithms and learns well from small datasets. It also performed well in run-time comparisons. The proposed algorithm thus saves time by automating quantitative phase analysis for future batches of PXRD data acquired in similar context. 


\end{document}